\documentclass[arxiv,article,accept,pdftex,moreauthors]{mdpi} 
\newcommand{\xadded}[1]{#1}
\newcommand{\xdeleted}[1]{}
\newcommand{\xreplaced}[2]{#1}

\firstpage{1} 
\pubvolume{0}
\issuenum{0}
\articlenumber{0}
\pubyear{2023}
\copyrightyear{2023}
\externaleditor{Academic Editors:}
\datereceived{2023} 
\dateaccepted{2023} 
\datepublished{2023} 
\hreflink{https://} % If needed use \linebreak
\usepackage{multirow,booktabs,bigstrut}
\usepackage[figuresright]{rotating}
\usepackage{changes}
\definechangesauthor[name={Author}, color=red]{V1.}
\definechangesauthor[name={Author}, color=blue]{V2.}
\newcommand{\tabincell}[2]{\begin{tabular}{@{}#1@{}}#2\end{tabular}} 
\Title{DKT-STDRL: Spatial and Temporal Representation Learning Enhanced Deep Knowledge Tracing for Learning Performance Prediction}

\TitleCitation{DKT-STDRL: Spatial and Temporal Representation Learning Enhanced Deep Knowledge Tracing for Learning Performance Prediction}

\Author{{Liting Lyu} %MDPI: Please carefully check the accuracy of names and affiliations. 
 $^{1}$\orcidA{}, Zhifeng Wang $^{2,}$*\orcidB{}, Haihong Yun $^{1}$, Zexue Yang $^{1}$ and Ya Li $^{1}$}

\AuthorNames{Liting Lyu, Zhifeng Wang, Haihong Yun, Zexue Yang and Ya Li}

\AuthorCitation{Lyu, L.; Wang, Z.; Yun, H.; Yang, Z.; Li, Y.}
\address{%
$^{1}$ \quad School of Computer Science and Technology, Heilongjiang Institute of Technology, Harbin 150050, China; lyuliting2022@163.com (L.L.); {yunhh126@126.com (H.Y.); yzx1978@126.com (Z.Y.); ly.218@163.com (Y.L.)} 
\\
$^{2}$ \quad School of Educational Information Technology, Central China Normal University, Wuhan 430079, China}

\corres{Correspondence: zfwang@ccnu.edu.cn}

\abstract{Knowledge tracing (KT) serves as a primary part of intelligent education systems. Most current KTs either rely on expert judgments or only exploit a single network structure, which affects the full expression of learning features. To adequately mine features of students’ learning process, Deep Knowledge Tracing Based on Spatial and Temporal Deep Representation Learning for Learning Performance Prediction (DKT-STDRL) is proposed in this paper. DKT-STDRL extracts spatial features from students’ learning history sequence, and then further extracts temporal features to extract deeper hidden information. Specifically, firstly, the DKT-STDRL model uses CNN to extract the spatial feature information of students’ exercise sequences. Then, the spatial features are connected with the original students’ exercise features as joint learning features. Then, the joint features are input into the BiLSTM part. Finally, the BiLSTM part extracts the temporal features from the joint learning features to obtain the prediction information of whether the students answer correctly at the next time step. Experiments on the public education datasets ASSISTment2009, ASSISTment2015, Synthetic-5, ASSISTchall, and Statics2011 prove that DKT-STDRL can achieve better prediction effects than DKT and CKT.}


\begin{document}

%%%%%%%%%%%%%%%%%%%%%%%%%%%%%%%%%%%%%%%%%%
% \setcounter{section}{-1} %% Remove this when starting to work on the template.
% \section{How to Use This Template}

% The template details the sections that can be used in a manuscript. Note that the order and names of article sections may differ from the requirements of the journal (e.g., the positioning of the Materials and Methods section). Please check the instructions on the authors' page of the journal to verify the correct order and names. For any questions, please contact the editorial office of the journal or support@mdpi.com. For LaTeX-related questions please contact latex@mdpi.com.%\endnote{This is an endnote.} % To use endnotes, please un-comment \printendnotes below (before References). Only journal Laws uses \footnote.

% The order of the section titles is: Introduction, Materials and Methods, Results, Discussion, Conclusions for these journals: aerospace,algorithms,antibodies,antioxidants,atmosphere,axioms,biomedicines,carbon,crystals,designs,diagnostics,environments,fermentation,fluids,forests,fractalfract,informatics,information,inventions,jfmk,jrfm,lubricants,neonatalscreening,neuroglia,particles,pharmaceutics,polymers,processes,technologies,viruses,vision

\section{Introduction}

At present, \xreplaced{learning management systems (LMSs),}{online education platforms} are widely welcomed~\cite{khlaif_covid-19_2021}. \xadded{LMSs are software systems for distance education based on the internet. LMSs have the functions of managing students and learning resources, providing students with the services of online learning, exercises, tests, communication, registration, scheduling, logging, and \mbox{so on~\cite{cavus_distance_2015}}. The typical LMSs can be divided into categories of open LMSs (e.g., EdX, \mbox{Moodle~\cite{pardos_adapting_2013})}, customized LMSs (e.g., ASSISTments~\cite{heffernan_future_2016}), and learning management ecosystems~\cite{gorshenin_toward_2018,cavus_distance_2015}.} \xreplaced{LMSs have }{The \xreplaced{LMS}{online education platform }has }rich resources, flexibility, and convenience, which brings new development opportunities for intelligent education~\cite{natella_systematic_2019,Lyu2022}. As a crucial research branch of intelligent education, knowledge tracing (KT), which promotes solving the problem of personalized tutoring for learners, has attracted more and more attention~\cite{teodorescu_continuous_2021}. KT can model and analyze the data of interactions from learners practicing online to obtain the potential law of the change of different students’ knowledge status, and then \xdeleted{to} predict students’ future learning performance~\cite{corbett_knowledge_1995}. 
\xadded{Specifically, knowledge tracing can model the students' practice process by logistic function, machine learning (such as hidden Markov models) or deep learning (such as recurrent neural networks, graph neural networks) algorithm models based on the students' practice records collected by \xreplaced{LMSs}{the LMS} such as ASSISTments and Coursera. Through the modeling and analysis of the interaction process between students and exercises, KT models can learn parameters about the hidden state of students' mastery of knowledge points (the probability of successfully mastering the corresponding knowledge points of exercises) from the explicit answer history sequence\xdeleted{,} and then can output the prediction of whether the student can answer the next question correctly or not. This is because it is generally believed in the field of education that exercises can test students' grasp of the knowledge points contained in the exercises, and students' grasp of the knowledge points can also be reflected through exercises.}
KT can \xreplaced{analyze}{analyse} and predict the learning status of different students, which is helpful for \xreplaced{LMSs}{online education platforms} to provide personalized learning programs and help students improve learning efficiency~\cite{rastrollo-guerrero_analyzing_2020}. When most \xdeleted{of} students can achieve better learning results, the educational level and social economy are expected to usher in further prosperity and development. As a result, more research has been conducted on KT.

Today, there are many KT models. First of all, Bayesian knowledge tracing (BKT) \xreplaced{was}{is} proposed in 1994~\cite{corbett_knowledge_1995}, which \xreplaced{took}{takes} whether students have mastered a certain knowledge point or not as a hidden variable and \xreplaced{used}{uses} a hidden Markov model (HMM) to model its changes in the learning process. BKT has laid an important foundation for the development of KT. For a long period of time, the research on KT \xreplaced{has mainly focused}{focuses} on the improvement and application of BKT, such as \xreplaced{Knowledge Tracing: Item Difficulty Effect Model (\mbox{KT-IDEM})}{KT-IDEM}~\cite{konstan_kt-idem_2011} and \xreplaced{Personalized Clustered BKT (PC-BKT)}{PC-BKT}~\cite{nedungadi_predicting_2014}. However, the traditional KT models represented by BKT \xreplaced{were}{are} limited by manual tags and \xreplaced{were}{are} limited by modeling and analyzing single concepts.  Then, deep knowledge tracing (DKT)~\cite{piech_deep_2015} \xreplaced{was}{is} put forward in 2015, which \xreplaced{applied}{applies} recurrent neural networks (RNNs) to modeling the learning process of students and predicting their future performance. DKT \xreplaced{solved}{solves} the problem of the discrete modeling of knowledge points without manually annotating the relationship between exercises and knowledge points.  Since then, more and more deep learning (DL) techniques \xreplaced{have been}{are} introduced into KT modeling tasks.  For example, dynamic key-value memory networks (DKVMNs)~\cite{zhang_dynamic_2017} \xreplaced{brought}{brings} in and  \xreplaced{improved}{improves} memory-augmented neural networks (MANNs) to automatically discover knowledge points in exercises and predict students’ knowledge state; convolutional neural networks (CNNs) \xreplaced{were}{are} introduced into CKT~\cite{shen_convolutional_2020} to analyze students’ personalized phased learning characteristics and predict learning performance. GKT~\cite{nakagawa_graph-based_2019} \xreplaced{built}{builds} a graph neural network (GNN) by using the correlation between knowledge points to predict learning performance. The KT models based on deep learning \xreplaced{had}{have} significant advantages in automatically extracting features, which \xreplaced{was}{is} suitable for mining potential rules from massive data without manual operation.  Compared with traditional KTs, KTs on the basis of DL \xreplaced{catered}{cater} to the development of big data in education. However, 
\xreplaced{the existing KT models had the problem that they expressed features inadequately. It can be found that most of the existing KTs on the basis of DL used relatively single network structures. For example, DKT just used RNN or LSTM to extract the temporal features of students' learning sequences, CKT just used CNN to extract the spatial features of students' learning sequences, and GKT just used GNN to learn the correlations between knowledge points. 
The feature extraction abilities of different DL network structures have different advantages and disadvantages. So, KTs based on single \xreplaced{networks}{network} had limitations in sufficiently extracting features, resulting in insufficiently mining and utilizing information based on original data and \xreplaced{a difficulty in obtaining a more \mbox{accurate prediction}.}{difficult to get more \mbox{accurate prediction}.}
}{most of the existing KT network structures are relatively simple, resulting in insufficient data feature extraction.}

To solve the problem that the KT model does not adequately express deep features, we propose Deep Knowledge Tracing Based on Spatial and Temporal Deep Representation Learning for Learning Performance Prediction (DKT-STDRL)\xadded{, which can fully explore and comprehensively utilize the spatial and temporal characteristics of students' exercise sequences, so as to obtain more accurate prediction results.} 
Inspired by DKT~\cite{piech_deep_2015} and CKT~\cite{shen_convolutional_2020}, this model combines the students’ exercise history and the spatial features as the joint learning features to further extract the temporal features of students’ learning process. Specifically, for a given sequence of students’ answer performance, we first extract the spatial features of students’ exercise sequences by using  multilayer convolution neural networks. \xadded{The spatial features can represent the learners' personalized learning efficiency, which can be reflected from the performance of continuous problem \mbox{solving~\cite{shen_convolutional_2020}}. }Then, we extract the temporal characteristics of learners’ learning performance sequences through bidirectional long short-term memory (BiLSTM)~\cite{graves_framewise_2005} based on the combination of spatial learning features and the original response performance. \xadded{The temporal features can represent the changing process of students' knowledge state.  }Because of the sufficient utilization of bidirectional sequence signals by BiLSTM~\cite{kim_gritnet_2018}, the DKT-STDRL considers both the past and future learning performance of students so that a better judgment of the current time can be obtained.

We sum up our main contributions as follows:

\begin{enumerate}
\item A new KT model is proposed by us, which is called Deep Knowledge Tracing Based on Spatial and Temporal Deep Representation Learning for Learning Performance Prediction (DKT-STDRL). This model tries to combine the advantages of CKT~\cite{shen_convolutional_2020} and DKT~\cite{piech_deep_2015} in extracting the spatial features and temporal features of students’ learning process, so as to mine students’ learning characteristics in two ways;
\item The DKT-STDRL model combines the spatial features of students learning sequence over the past period of time extracted by the multilayer convolutional neural network with the data of students’ historical answers to obtain the joint learning features. Then, we employ BiLSTM to further learn the time-series information of the joint learning features. The overlapped neural networks make the model take into account the spatial learning features when extracting the temporal features of students learning, so it is conducive to mining and obtaining deeper learning information of students;
\item We use BiLSTM to perform a two-way time series analysis of the joint learning features, which enables the model to analyze the learning features of students at each time step from both a past and future perspective, so as to obtain more accurate predictions;
\item We \xreplaced{conducted}{have done} sufficient experiments on five commonly used public educational datasets to demonstrate that DKT-STDRL obviously outperforms the CKT\xadded{~\cite{shen_convolutional_2020}} and DKT\xadded{~\cite{piech_deep_2015}} models on ACC, AUC, $r^{2}$, and RMSE;
\item We \xreplaced{completed}{have done} enough experiments to compare the prediction performance of DKT-STDRL with DKT\xadded{~\cite{piech_deep_2015}}, CKT\xadded{~\cite{shen_convolutional_2020}}, and four variants of DKT-STDRL (DKT-TDRL, DKT-SDRL1, DKT-STDRRP, and DKT-STDRRJ) when they are set {up with the} %Please check that the intended meaning is retained.
% We confirm.
 same hyperparameters on the same datasets, to observe the impacts of the different aspects of spatial features, temporal features, prior features, and joint features on the prediction metrics.
\end{enumerate}

The rest of the paper is organized as follows: first, Section~\ref{sec2} provides a brief overview of the research work on KT and the research motivations. Then, in Section~\ref{sec3}, we formally define the problem to be solved by the KT task, the implementation details of the components of the DKT-STDRL model structure, and the loss function. Section~\ref{sec4} compares the experimental results of DKT-STDRL with CKT, DKT, and the variants of the DKT-STDRL model on the same five open datasets. At last, the conclusion of this paper and prospective research work are pointed out in Section~\ref{sec5}.

%%%%%%%%%%%%%%%%%%%%%%%%%%%%%%%%%%%%%%%%%%
\section{Related Work and Motivation}\label{sec2}

The work of KT is partitioned into two phases in general: the development of KT models on the basis of traditional research methods and the development of KT models on the basis of deep learning.

The first stage of KT development is from 1994 to around 2015. During this period, the research on KT \xreplaced{was}{is} mainly based on traditional research methods, such as probability maps and psychological theories. Among them, probability map-based models represented by Bayesian knowledge tracing (BKT)~\cite{corbett_knowledge_1995} \xreplaced{were}{are} the main part. BKT \xreplaced{took}{takes} the knowledge state of whether students \xreplaced{mastered}{master} a knowledge point or not as a hidden variable, the result of whether they \xreplaced{answered}{answer} correctly the corresponding questions of the knowledge point or not as an observation variable, and \xreplaced{used}{uses} the hidden Markov model to represent the changes of the skills status of learners, and then \xreplaced{predicted}{predicts} the probability that learners \xreplaced{acquired}{acquire} the skill point. BKT \xreplaced{solved}{solves} the problem of KT for the first time. Subsequent research has also produced some variants of BKT that \xreplaced{improved}{improve} the performance of BKT in different ways. For example, C. Carmona et al. \xreplaced{proposed}{propose} introducing a layer structure to represent the prerequisite relationships in actual educational scenarios, consequently acquiring a more accurate model used for predicting students’ knowledge status~\cite{carmona_introducing_2005}. KT-IDEM~\cite{konstan_kt-idem_2011} \xreplaced{was}{is} proposed to more accurately predict students’ practice performance by adding the dependence of item difficulty to the probability map. BKT and its variants \xreplaced{had}{have} simple structures and strong explanatory ability. During this period, BKT \xreplaced{was}{is} of great significance in constructing intelligent tutorial systems and \xreplaced{learning management systems}{the online education platforms}\xadded{, such as the ACT Programming Tutor~\cite{corbett_knowledge_1995} and edX~\cite{pardos_adapting_2013}}. \xadded{For example, BKT improved the prediction of the test performance on the ACT Programming Tutor of students  \xreplaced{through modeling}{by using Bayesian to estimate} the change of the probability that the student grasped each rule~\cite{corbett_knowledge_1995}. EdX used variants of BKT to acquire students' practice performance on several questions in the past period, modeling students' mastery of knowledge components and, thus, predicting students' test performance in a later period.  This helps to analyze the learning effect of students on the edX platform~\cite{pardos_adapting_2013}.  }
However, BKT and its variants \xreplaced{ignored}{ignore} the long-term temporal dependence of students learning process because they \xreplaced{were}{are} based on a Markov chain structure. Furthermore, BKT \xreplaced{was}{is} not suitable for future scenarios with large amounts of data, because it \xreplaced{relied}{relies} on labeling the data manually. In addition, some scholars have applied psychological theory to KT, such as item response theory (IRT)~\cite{drasgow_item_1990}, DINA~\cite{de_la_torre_dina_2009}, and so on. Although models based on psychometric measurement theory \xreplaced{could}{can} be well interpreted, the simple parameter settings  \xreplaced{limited}{limit} the models’ ability to encode complex features. Therefore, though a number of effective models emerged in the first stage of KT, the model \xreplaced{structures were}{structure is} too simple to express more complex features. Moreover, there  \xreplaced{was}{is} a bottleneck in the development of data processing due to the need for \mbox{manual operation}. 

The second stage of KT is from 2015 to the present. Inspired by the great success of deep learning in the field of speech processing \cite{Wang2022t, Zeng2018, Wang2021m, Zeng2020, Wang2020h, Zeng2021b, Wang2018a, Zeng2022a, Wang2015b} and computer vision \cite{Wang2022ac, Zeng2020a, Wang2021, Zeng2021c, Wang2017, Zeng2022, Li2023, Wang2015a, Zeng2022b}, deep learning was introduced into KT. In 2015, deep knowledge tracing (DKT)~\cite{piech_deep_2015} \xreplaced{was}{is} proposed, marking the beginning of an era in which DL technologies \xreplaced{drove}{drive} the evolution of KT. DKT \xreplaced{used}{uses} long short-term memory (LSTM) to learn the procedure that the learners’ skill status \xreplaced{changed}{change} and \xreplaced{made}{make} a prediction about learners’ future exercise performance. Compared with BKT, the LSTM model used by DKT \xreplaced{could}{can} make use of long-term time series dependence, which \xreplaced{conformed to the long-term dependence of the state of knowledge at each moment in the actual learning process}{conforms more to people’s real studying law} and \xreplaced{helped}{helps} to obtain more accurate prediction results. In addition, DKT \xreplaced{could}{can} automatically extract features without labeling them, which \xreplaced{saved}{saves} labor costs and \xreplaced{avoided}{avoids} human errors. However, DKT only \xreplaced{considered}{considers} whether students answered correctly or not, ignoring other characteristics of the learning process. Liang Zhang et al. \xreplaced{supplemented}{supplement} other relevant features, (such as time, reminder utilization, and {attempt} %Please check that the intended meaning is retained. 
 counts) collected by education information platforms to the DKT model and \xreplaced{acquired}{acquire} more accurate prediction outputs~\cite{zhang_incorporating_2017}. Although the model \xreplaced{could}{can} predict better by adding the exercise performance information in other aspects, these features \xreplaced{were}{are} difficult to fully express the student’s personalized learning features. Shuanghong Shen et al. \xreplaced{proposed}{propose} convolutional knowledge tracing (CKT)~\cite{shen_convolutional_2020}, which \xreplaced{used}{uses} convolutional neural networks (CNNs) to extract students’ personalized prior knowledge and learning rates from their answer histories. Although extracting and utilizing personalized learning features \xreplaced{could}{can} improve prediction accuracy, characterizing temporal changes \xreplaced{was}{is} not sufficient because only a gate linear unit (GLU)~\cite{dauphin_language_2017} \xreplaced{was}{is} used to control the forward temporal dependence of personalized prior knowledge. GritNet~\cite{kim_gritnet_2018} bidirectionally \xreplaced{analyzed}{analyzes} students’ exercise records by BiLSTM to predict students’ future performance, which fully \xreplaced{extracted}{extracts} the temporal features of the learning process\xadded{, namely, the knowledge state for each knowledge point during each practice interaction}.\xadded{ In the whole learning process, temporal factors such as {remebering} %Please check that the intended meaning is retained.
 	 and forgetting could affect the students' mastery of knowledge at set intervals. Neural networks could learn representation features of the potential knowledge mastery state under the influence of these temporal factors. } However, it \xreplaced{did}{does} not take into account other learning features\xadded{, such as personalized prior knowledge features or individualized learning rate features}. Moreover, the experimental dataset, Udacity data, \xreplaced{was}{is} not commonly used. Moreover, GritNet \xreplaced{was}{is} only proved to be superior to the standard logistic-regression-based method. The dynamic key-value memory network (DKVMN)~\cite{zhang_dynamic_2017} \xreplaced{employed}{employs} the idea of memory-augmented neural networks (MANNs), which \xreplaced{recorded}{records} and \xreplaced{updated}{updates} the knowledge points and students’ mastery of knowledge with two matrices, respectively. DKVMN \xreplaced{had}{has} an advantage in explanation. GKT~\cite{nakagawa_graph-based_2019} \xreplaced{built}{builds} a graph neural network (GNN)~\cite{gori_new_2005} on the basis of the association between knowledge points, which \xreplaced{could}{can} help improve prediction and interpretation performance~\cite{nakagawa_graph-based_2019}. Sequential key-value memory networks (SKVMN)~\cite{abdelrahman_knowledge_2019} \xreplaced{employed}{employs} the triangular membership function to diagnose students’ mastery of knowledge concepts and \xreplaced{used}{uses} the improved LSTM to capture the features of the learning process, which \xreplaced{showed}{shows} good performance in prediction accuracy and explanation. Qi Liu et al. \xreplaced{attempted}{attempt} to combine the students’ exercise performance with the features extracted from the test content, and proposed an exercise-enhanced recurrent neural network (EERNN) and EKT~\cite{liu_ekt_2021}, which \xreplaced{exploited}{exploit} BiLSTM, a Markov property, or an attention mechanism. The experiment proved that supplementing the feature information of the problem content \xreplaced{could}{can} help advance the prediction effect. The self-attention mechanism \xreplaced{was}{is} utilized in self-attentive knowledge tracing (SAKT)~\cite{pandey_self-attentive_2019} to model learners’ performance on related questions, and then \xreplaced{to predict}{predicts} learners’ future responses. It \xreplaced{solved}{solves} the sparse dataset problem and \xreplaced{achieved}{achieves} good prediction results. Although previous research has made up for the deficiencies in different aspects of KT models and promoted the development of KT based on deep learning, most models \xreplaced{had}{have} relatively simple feature extraction methods, leading to the inadequate use of information on students’ practice records. \xadded{A comparison of previous main research in KT \xreplaced{is shown }{shows as }in Table~\ref{tab6}.}

As far as we \xreplaced{knew}{know}, there \xreplaced{were}{are} spatial features and temporal features in students learning sequences\xadded{, which had location correlations and time correlations}. 
\xadded{The spatial features can be extracted by CNN.  Typical CNN~\cite{lecun_gradient-based_1998} structures were mainly composed of convolution layers and pooling \xreplaced{layers}{layer} alternately, among which the core was the convolution layer.  The convolution layer contained multiple convolution kernels.  The convolution kernel contained multiple parameters.  By translating the convolution kernel to scan different positions of samples, parameters in the convolution kernel could learn some local features in the sample space, which was conducive to the judgment of classification.  \xreplaced{The pooling}{Pooling} layer played the role of sub-sampling to reduce model parameters and reduce model complexity.  CNN was often used for two-dimensional image processing, such as handwriting recognition~\cite{lecun_gradient-based_1998}, due to its good spatial feature extraction ability.  One-dimensional sequence samples could be regarded as special two-dimensional structures, so CNN was also suitable for processing \xreplaced{time-series}{time series} data, such as natural language processing~\cite{gu_recent_2015}, and only one-dimensional convolution \xreplaced{kernels were}{kernel is} required.  In KT, we used one-dimensional convolution to extract the spatial features of students' practice sequences. In addition, temporal features can be extracted by BiLSTM~\cite{graves_framewise_2005}.  BiLSTM is a method based on LSTM~\cite{hochreiter_long_1997} for bidirectional parameter learning, which consists of circularly connected memory modules.  Each module contains three gate units, which are used to control the input, output\xadded{,} and forgetting operations of transmitted information.  Through this gating mechanism, the long-term dependence of \xadded{the }information transfer process was solved, which was conducive to \xadded{the }time series analysis of samples.  LSTM and BiLSTM were often used in sequence learning tasks~\cite{graves_framewise_2005}, such as \xdeleted{the }word sequence processing.  In KT, we used BiLSTM to extract the temporal features of student exercise sequences. }
Therefore, we \xreplaced{tried}{try} to integrate CNN and BiLSTM to extract the spatial features and temporal features of interaction sequences, which \xreplaced{helped}{helps} to use feature information fully~\cite{zeng_spatial_2021}. In addition, the BiLSTM further \xreplaced{improved}{improves} the adequate utilization of the information by extracting features in both directions.

\startlandscape

\begin{table}[H]
\footnotesize
\vspace{6pt}
	%\textcolor{blue}{
	%\centering
	\caption{Comparison of previous main research in KT. The ``Category’’ column represents the two main types of knowledge tracing methods, the ``KTs’’ column stands for the different knowledge tracing models, the ``Year’’ column is the year that the knowledge tracing model was firstly published, the ``Technology’’ column means the main machine learning or deep learning techniques used, the ``Key Article’’ column shows the key articles involved in the method, the ``Utility’’ column serves as the educational application scenarios, and the ``Characteristics’’ column summarizes the main features of the knowledge tracing method.}
	\begin{tabular}{m{2.6cm}llm{2cm}m{1cm}m{3cm}m{6cm}m{7.45cm}}
		\toprule
		\textbf{{Category} 
} & \textbf{KTs} & \textbf{Year} & \textbf{Technology} & \textbf{AUC} & \textbf{Key Article} & \textbf{Utility} & \textbf{Characteristics}\\
		\midrule
		\multirow{3}{*}{\tabincell{l}{{Probability Graph}\\{based KT}}} %MDPI: PLEASE confirm if ``Probability Graph'' should be merged with `` based KT'' in one row
		%Authors: We confirm that ``Probability Graph'' should be merged with `` based KT'' in one row.
   & BKT   & 1994  & HMM   & 0.670  & Corbett et al.~\cite{corbett_knowledge_1995} & Programming performance prediction & Single knowledge \\
		\cmidrule{2-8}     &KT-IDEM & 2011  & HMM+IRT & 0.690  & Pardos et al.~\cite{konstan_kt-idem_2011} & Performance prediction on math exercises & Ignore long term dependences \\
		\midrule
		\multirow{10.5}{*}{\tabincell{l}{{Deep Learning}\\ {based KT}}}& DKT   & 2015  & RNN/LSTM & 0.805 & Piech et al.~\cite{piech_deep_2015} & Performance prediction on math courses & Multiple knowledge, single feature \\
		\cmidrule{2-8}          & DKT+  & 2017  & RNN/LSTM & 0.863 & Zhang et al.~\cite{zhang_incorporating_2017} & Performance prediction on {statistics} %Please check that the intended meaning is retained.
		%Authors: The intended meaning is retained.
		 courses & Multiple knowledge, multiple features \\
		\cmidrule{2-8}          & DKVMN & 2017  & MVNN  & 0.816 & Zhang et al.~\cite{zhang_dynamic_2017} & Performance prediction on math exercises & Adaptively update knowledge mastery \\
		\cmidrule{2-8}    & SKVMN & 2019  & LSTM+MVNN & 0.836 & Abdelrahman et al.~\cite{abdelrahman_knowledge_2019} & Performance prediction on scientific courses & Multiple knowledge, long-term dependencies \\
		\cmidrule{2-8}    & GKT   & 2019  & GNN   & 0.723 & Nakagawa et al.~\cite{nakagawa_graph-based_2019} & Performance prediction on math courses & Model relationship between exercises and knowledge \\
		%Authors: We confirm that ``Deep Learning'' should be merged with `` based KT'' in one row.
		\cmidrule{2-8}          & EKT   & 2019  & LSTM  & 0.850  & Liu et al.~\cite{liu_ekt_2021} & Performance prediction on math courses & Apply semantic information of exercises \\
		\cmidrule{2-8}          & SAKT  & 2019  & FFN+MSA & 0.848 & Pandey et al.~\cite{pandey_self-attentive_2019} & Performance prediction on scientific courses & Model relationship among knowledge points \\
		\cmidrule{2-8}          & CKT   & 2020  & CNN   & 0.825 & Shen et al.~\cite{shen_convolutional_2020} & Performance prediction on math exercises & Extract spatial features \\
		\bottomrule
	\end{tabular}%
	\label{tab6}%
%	}
\end{table}%
\finishlandscape

%%%%%%%%%%%%%%%%%%%%%%%%%%%%%%%%%%%%%%%%%%
\vspace{6pt}
\section{Proposed Method}\label{sec3}
\subsection{Problem Definition}
The function of KT is described as follows: for a given dataset of students’ exercise performance history, the interaction sequence when a student answers $l$ times can be {expressed as} %MDPI: Please confirm if the bold of all symbol in this paper should be retained. If so, please keep the same variable in the same format. If not, please remove it.
%Authors: We confirm the bold of all symbol in this paper should be retained.
 $\boldsymbol{{U}}_{l}=\left(\boldsymbol{u}_{1}, \boldsymbol{u}_{2}, \boldsymbol{u}_{3}, \ldots, \boldsymbol{u}_{t}, \ldots \boldsymbol{u}_{l}\right)$, where $l$ is the length of the sequence. \mbox{$\boldsymbol{u}_{t}=\left(s_{t}, r_{t}\right)$} represents the exercise record of the student at time step $t$, $s_{t}$ is the number of the question at time step $t$, and $r_{t}$ represents the corresponding answer result. $r_{t}$ has only two values: 0 or 1. When $r_{t}=1$, the student answered the question correctly at the corresponding time step. When $r_{t}=0$, the student answered the question incorrectly at the corresponding time step. KT uses $\boldsymbol{U}_{l}$  to learn the students’ skill status, for predicting whether the students can successfully solve the problem $s_{t+1}$ at each next time step $t+1$, namely, $r_{t+1}$.
\subsection{Model Architecture}
\begin{figure}[H]
\includegraphics[width=0.8\textwidth]{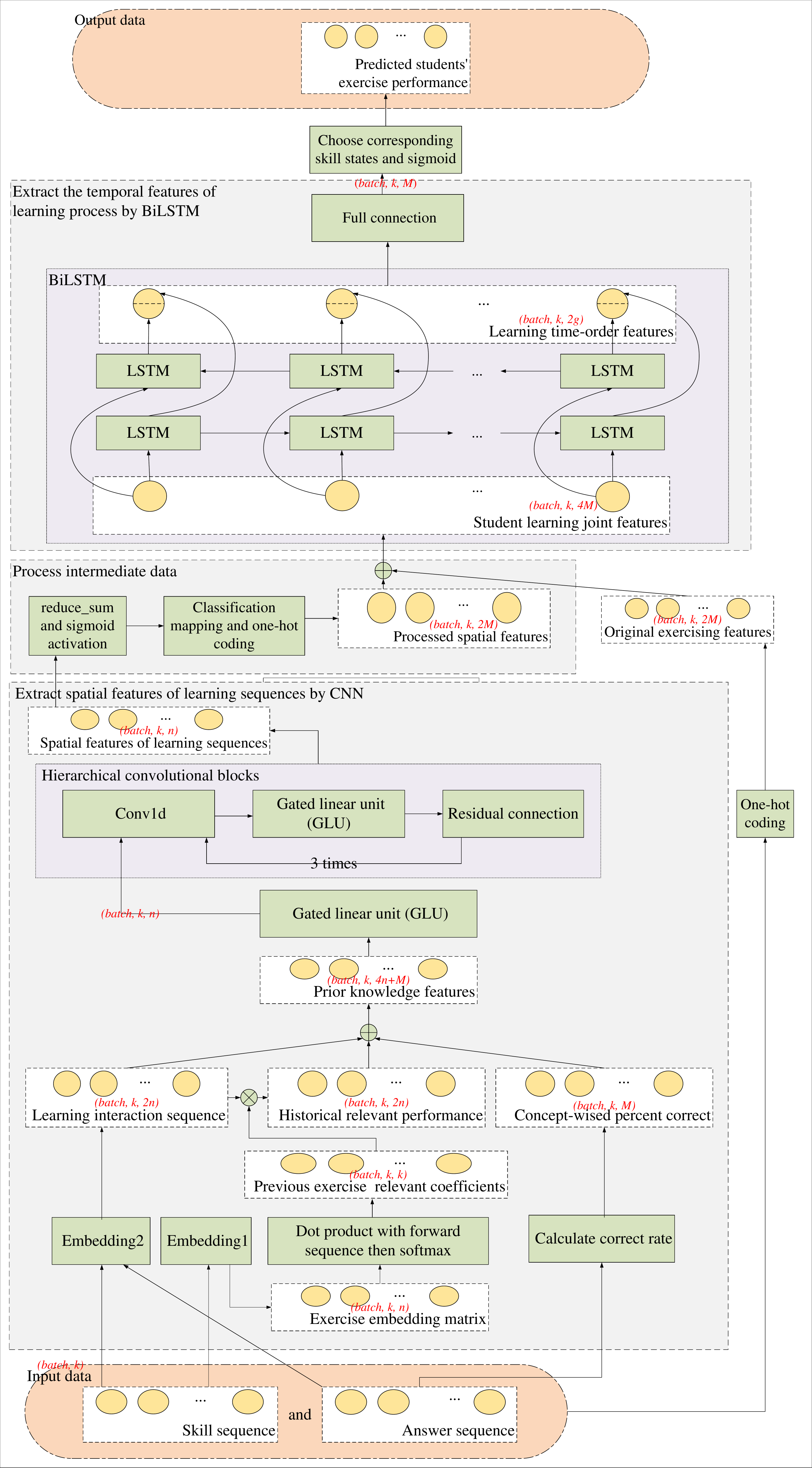}
\caption{The architecture of\xadded{ the} DKT-STDRL model. The red text in the figure \xreplaced{shows}{show} the shape of the data, where\xdeleted{,} {$batch$} %MDPI: please check whether the italic should be retained. If so, please unify the format in the figure. If not, please remove it.
%Authors: The italic should be retained. so, we have unified the format in the figure.
 equals the \xdeleted{to }batch size. {$k$}  
 represents the max sequence length. {$n$}  
 is the number of dimensions of the embedding matrix.{ $M$}  
 is the \xreplaced{count}{counts} of different skills. $g$ is the number of units of each LSTM module.\label{fig1}}
\end{figure}

Deep Knowledge Tracing Based on Spatial and Temporal Deep Representation Learning for Learning Performance Prediction (DKT-STDRL) consists of 3 parts: the part of extracting the spatial features of students’ learning sequences by CNN, the part of intermediate data processing\xadded{,}  and the part of extracting students’ temporal features in the learning process by BiLSTM. 
\xadded{
In the part of spatial feature extraction, a multi-layer convolutional structure~\cite{lecun_gradient-based_1998} is used to obtain students' personalized learning efficiency information as supplementary information.  The goal of the intermediate data processing part is to combine the students' practice history and personalized learning efficiency information for the subsequent sequence feature extraction part.  Therefore, the intermediate data processing part merges the one-hot coded spatial features and the exercise history features into the joint features of student learning, and then passes them to the next part.  The next part, namely, temporal feature extraction, adopts the BiLSTM~\cite{graves_framewise_2005} structure to further extract bidirectional time-sequence features from the student learning joint features, so as to obtain the information about the change of students' knowledge state in the learning process and then predict students' performance in the next time slice. 
}
The architecture diagram of DKT-STDRL is shown in Figure~\ref{fig1}. 

\xadded{
\xreplaced{As shown in }{From the }Figure~\ref{fig1}, it can be noted that the DKT-STDRL improves the model structure of DKT~\cite{piech_deep_2015} and CKT~\cite{shen_convolutional_2020}.
}

\xadded{
DKT directly used LSTM~\cite{hochreiter_long_1997} to extract the temporal features (hidden knowledge state of students) of the sequences from the students' practice history, and then output the prediction results of the next time slice~\cite{piech_deep_2015}.  Compared with DKT, DKT-STDRL is improved in two aspects.
First, the input information of \xadded{the }sequential learning structure is more abundant.  The information data extracted from sequence features is no longer only the original student practice records, but also includes the spatial features extracted from the student practice sequences using CNN~\cite{lecun_gradient-based_1998}. In this way, the spatial features can be regarded as abstract characteristics of students' personalized learning efficiency.  Spatial features are added as the input of\xadded{ the} temporal feature extraction structure for analyzing and predicting students' practice sequences from the spatial perspective before analyzing students' practice sequences from the temporal perspective.  Thus, the learning analysis process of the model is more comprehensive, because the change of knowledge states of students in the learning process and the influencing factors of students' personalized learning efficiency are analyzed at the same time, and then\xadded{ a} more accurate prediction can be obtained.  
Secondly, the learning ability of the structure for temporal features learning is improved.  The temporal features learning structure is changed from LSTM~\cite{hochreiter_long_1997} to BiLSTM~\cite{graves_framewise_2005}, which enables the model to learn not only the forward sequence features, but also the reverse sequence features.  The two-way learning mode enhances the ability of the model to adjust internal parameters so that it is easy to obtain more accurate prediction results.  The extraction method of bidirectional temporal features is in line with the practical significance, which enables the KT model to consider students' future performance as well as their past performance, which enables more accurate judgments to be obtained when analyzing students' knowledge mastery at each time step.
}

%%\unskip

\xadded{
CKT~\cite{shen_convolutional_2020} extracted the characteristics of personalized prior knowledge from students' practice records, and then extracted the characteristics of the learning rate based on the characteristics of the personalized prior knowledge using CNN~\cite{lecun_gradient-based_1998}, then outputting the prediction.  Although the spatial features of student practice sequences are extracted, CKT has defects in temporal \xreplaced{feature}{features} extraction.  Compared with CKT~\cite{shen_convolutional_2020}, DKT-STDRL enhanced the time series feature extraction capability of CKT.  After simple processing, the spatial features extracted from sequences are input into\xadded{ 
the} BiLSTM structure to further extract the bidirectional temporal features.  In this way, the prediction output of the model is based on the comprehensive analysis of spatial features and temporal features, which is to say that our model can simultaneously analyze the personalized learning rate of students and the change of knowledge state of students in the learning process, and it is thereby easy to obtain more accurate prediction results. 
}

\subsubsection{Using {CNN} to Extract Spatial Features of Students’ Learning Sequences} %ATTN: Title altered.
%Authors: We confirm.
Because CKT has significant advantages in modeling students’ personalized learning interaction process~\cite{shen_convolutional_2020}, DKT-STDRL borrows the CKT model structure. Different from CKT, this component of the DKT-STDRL model needs only to extract the spatial features of students’ learning sequences without predicting the performance of the next time step. The steps are as follows:
\begin{enumerate}
\item	Convert the input to an embedded matrix. \\
In order to make the model better extract spatial features of students' learning process, CKT uses the embedding matrix to represent the given students’ exercise history $\boldsymbol{U}_{k}$. $k$ is the max sequence length. Specifically, ${s}_{t}$ is randomly initialized as an n-dimensional vector $\boldsymbol{s}_{t}$ by embedding; $n$ is far less than the total counts of knowledge points in the whole dataset $M$. Then, according to whether the answer result is correct or not, the same n-dimensional zero vector is spliced on the right or left side of $\boldsymbol{s}_{t}$, so that $\boldsymbol{U}_{k}$  is transformed into a $k \times 2n$ dimensional matrix through embedding~\cite{shen_convolutional_2020};
\item	Calculate the prior knowledge of different students.\\
Because the learners’ prior knowledge \xreplaced{varies }{vary }with each individual and can be reflected by students’ past exercise performance and correct ratio on relevant skills~\cite{wang_neural_2020}, CKT obtains the prior knowledge of different students by calculating {values for} %Please check that the intended meaning is retained.
 historical relevant performance (HRP) and concept-wise percent correct (CPC)~\cite{shen_convolutional_2020};
\item	Calculate the learning rate of different students.\\
Because the learning efficiency of different students is different, and the continuous exercise performance of students over a period can reflect the learning efficiency of different students in different learning stages, CKT extracts the learning rate of different students through multi-layer CNN.
\end{enumerate}

{Through the} %MDPI: indent has been added, please confirm
%Authors: We confirm
 above process, the spatial learning features of \xreplaced{students' learning sequences }{ the knowledge status changing process of students }can be obtained. The process is visualized by the formula as follows:
\begin{equation}
\begin{aligned}
&\boldsymbol{e}_{t}= \begin{cases}{\left[\boldsymbol{s}_{t} \oplus \mathbf{0}\right],} & \text { if } r_{t}=1 \\ {\left[\mathbf{0} \oplus \boldsymbol{s}_{t}\right],} & \text { if } r_{t}=0\end{cases}\\
\end{aligned}
\end{equation}
\begin{equation}
\begin{aligned}
&\boldsymbol{F}_{L I S}=\left(\boldsymbol{e}_{1}, \boldsymbol{e}_{2}, \ldots, \boldsymbol{e}_{k}\right)^{\top}\\
\end{aligned}
\end{equation}
\begin{equation}
\begin{aligned}
&\left\{\begin{array}{l}
\boldsymbol { relation } t(j)=\text { Masking }\left(\boldsymbol{s}_{j} \cdot \boldsymbol{s}_{t}\right), j \in(t, k) \\
\boldsymbol { weight }_{t}(j)=\text { Softmax }\left(\boldsymbol { relation }_{t}(j)\right), j \in(1, k)
\end{array}\right.\\
\end{aligned}
\end{equation}
\begin{equation}
\begin{aligned}
&\boldsymbol{F}_{H R P_{t}}(t)=\sum_{j=1}^{t-1} \boldsymbol { weight }{ }_{t}(j) \boldsymbol{e}_{j}\\
\end{aligned}
\end{equation}
\begin{equation}
\begin{aligned}
\boldsymbol{F}_{C P C_{t}}(m)=\frac{\sum_{j=0}^{t-1} r_{j}^{m}==1}{\operatorname{total}\left(s^{m}\right)}
\end{aligned}
\end{equation}
\begin{equation}
\begin{aligned}
\left\{\begin{array}{l}
\boldsymbol{F}_{J P F}=\boldsymbol{F}_{L I S} \oplus \boldsymbol{F}_{H R P} \oplus \boldsymbol{F}_{C P C} \\
\boldsymbol{F}_{I L A}=\left(\boldsymbol{F}_{J P F} * \boldsymbol{W}_{3}+\boldsymbol{b}_{3}\right) \otimes \sigma\left(\boldsymbol{F}_{J P F} * \boldsymbol{W}_{4}+\boldsymbol{b}_{4}\right)
\end{array}\right.
\end{aligned}
\end{equation}
\begin{equation}
\begin{aligned}
&\boldsymbol{F}_{K S}=\left\{G L U\left(\operatorname{Conv} 1 d\left(\boldsymbol{F}_{I L A}\right)\right)\right\}_{3}
\end{aligned}
\end{equation}
where $\boldsymbol{F}_{L I S}$ is the embedded matrix expression of the learning interaction sequence $\boldsymbol{U}_{k}$. $Masking$ is an operation used to exclude subsequences. $\boldsymbol{weight}$ is used to estimate the resemblance between the present question and the previous question, and to then comprehensively analyze the impact of previous exercise performance on the current exercise. $\boldsymbol{F}_{H R P}$  is a historically related performance feature. $m$ is the number of the knowledge point. $\boldsymbol{F}_{C P C_{t}}(m)$ indicates the correct rate when the student answers the question of a knowledge point   at a certain time. $\boldsymbol{F}_{J P F}$ is the prior knowledge feature obtained by connecting $\boldsymbol{F}_{L I S}$, $\boldsymbol{F}_{H R P}$,  and $\boldsymbol{F}_{C P C}$. The spatial features of people’s studying processes  $\boldsymbol{F}_{K S}$  can be obtained from $\boldsymbol{F}_{J P F}$ through 3 layers of nonlinear transformations of a gated linear unit (GLU) and one-dimensional convolution operation~\cite{shen_convolutional_2020}. 

\subsubsection{Intermediate Data Processing}
The main goal of intermediate data processing is to take the spatial features extracted from the previous part as supplementary information \xreplaced{regarding}{of} students’ original exercise features, and then input them into the next part to extract the temporal features of students’ learning process. Specifically, firstly, the spatial learning features $\boldsymbol{F}_{K S}$ \xreplaced{are}{is} activated by the sigmoid function. Then, the processed spatial features are suitable for the next part after classification and one-hot coding. Finally, the spatial features are combined with the student learning record sequences processed by one-hot coding to form joint features as the input of BiLSTM. The process is shown by the formula as follows:
\begin{equation}
\begin{aligned}
\boldsymbol{F}_{K S}^{\prime}=\sigma\left(\boldsymbol{F}_{K S}\right) \\
\end{aligned}
\end{equation}
\begin{equation}
\begin{aligned}
\boldsymbol{F}_{K S}^{\prime \prime}(t)= \begin{cases}1, & \text { if } \boldsymbol{F}_{K S}^{\prime}>0.5 \\
0, & \text { if } \boldsymbol{F}_{K S}^{\prime} \leq 0.5\end{cases} \\
\end{aligned}
\end{equation}
\begin{equation}
\begin{aligned}
\boldsymbol{A} \boldsymbol{P}_{1}(t)=\boldsymbol{F}_{K S}^{\prime \prime}(t) \cdot M+s_{t} \\
\end{aligned}
\end{equation}
\begin{equation}
\begin{aligned}
\boldsymbol{F}_{I L F}=O n e H o t\left(\boldsymbol{A} \boldsymbol{P}_{1}, 2 M\right) \\
\end{aligned}
\end{equation}
\begin{equation}
\begin{aligned}
\boldsymbol{A} \boldsymbol{P}_{2}(t)=\left(1-r_{t}\right) \cdot M+s_{t} \\
\end{aligned}
\end{equation}
\begin{equation}
\begin{aligned}
\boldsymbol{F}_{L R S}=O n e H o t\left(\boldsymbol{A P}_{2}, 2 M\right) \\
\end{aligned}
\end{equation}
\begin{equation}
\begin{aligned}
\boldsymbol{F}_{L J F}(t)=\boldsymbol{F}_{I L S}(t) \oplus \boldsymbol{F}_{L R S}(t)
\end{aligned}
\end{equation}
where $\boldsymbol{A P}_{1}$ and $\boldsymbol{A P}_{2}$ are used to calculate the position of 1, which {is} %Please check that the intended meaning is retained.
% We confirm.
 used for the one-hot coding transformation of the spatial features and learning record sequence.   $\boldsymbol{F}_{I L F}$ represents the spatial features encoded by one-hot, and $\boldsymbol{F}_{L R S}$ represents the learning record sequences encoded by one-hot. The two are connected and combined to form the joint feature of students’ learning $\boldsymbol{F}_{L J F}$. $\boldsymbol{F}_{L J F}$ is a $k \times 4M$ matrix.

\subsubsection{Using {BiLSTM} to Extract the Temporal Features of Students’ Learning Process} %ATTN: Title altered.
%Authors: We confirm.
Inspired by GritNet~\cite{kim_gritnet_2018}, this part of the model introduces BiLSTM to extract the temporal features of students’ studying \xreplaced{procedures}{procedure} and make \xadded{a }prediction about learners’ recent learning effect. Because BiLSTM can pass information forward and backward at the same time, the DKT model with \xadded{the }BiLSTM structure can make use of students’ past exercise performance information and future exercise performance information at the same time. Compared with the original DKT consisting of the LSTM structure, which can only use past information, the DKT based on the BiLSTM structure can use more sufficient information to adjust the parameters of the model, so as to obtain more accurate judgment results. The processing process of intermediate data in this part is as follows:
\begin{equation}
\begin{aligned}
&\overrightarrow{\boldsymbol{h}_{t}}=\operatorname{LSTM}\left(\boldsymbol{F}_{L J F}(t), \overrightarrow{\boldsymbol{h}_{t-1}}\right) \\
\end{aligned}
\end{equation}
\begin{equation}
\begin{aligned}
&\overleftarrow{\boldsymbol{h}_{t}}=\operatorname{LSTM}\left(\boldsymbol{F}_{L J F}(t), \overleftarrow{\boldsymbol{h}_{t+1}}\right) \\
\end{aligned}
\end{equation}
\begin{equation}
\begin{aligned}
&\boldsymbol{h}_{t}=\overrightarrow{\boldsymbol{h}_{t}} \oplus \overleftarrow{\boldsymbol{h}_{t}} \\
\end{aligned}
\end{equation}
\begin{equation}
\begin{aligned}
&\boldsymbol{h}_{t}^{\prime}=\boldsymbol{h}_{t} * \boldsymbol{W}_{5}+\boldsymbol{b}_{5}
\end{aligned}
\end{equation}
where $\overrightarrow{\boldsymbol{h}_{t}}$ is the hidden state in chronological order when it is time step $t$. $\overleftarrow{\boldsymbol{h}_{t}}$ is the hidden state in reverse time order when it is time step $t$. $\overrightarrow{\boldsymbol{h}_{t}}$  and $\overleftarrow{\boldsymbol{h}_{t}}$  are combined to obtain $\boldsymbol{h}_{t}$, and $\boldsymbol{h}_{t}$ is the temporal features of students’ learning process extracted by BiLSTM. $\boldsymbol{h}_{t}^{\prime}$  can be regarded as the hidden knowledge state after comprehensive analysis. Through $\boldsymbol{h}_{t}^{\prime}$  and an exercise sequence, the exercise performance at the next time step $p_{t+1}$  can be easily obtained by simply finding the knowledge state of the corresponding problem and being dealt with by the sigmoid function.

\subsection{Optimization}
In order to optimize the model, the loss function of cross-entropy and the Adam optimizer are adopted. The loss function is as follows:
\begin{equation}
\text { Loss }=-\sum_{i=1}^{l}\left(r_{i} \log p_{i}+\left(1-r_{i}\right) \log \left(1-p_{i}\right)\right)
\end{equation}

%%%%%%%%%%%%%%%%%%%%%%%%%%%%%%%%%%%%%%%%%%
\section{Results and Discussion}\label{sec4}
\subsection{Experimental Datasets}
For the purpose of fair comparison with other KTs, experiments \xreplaced{were}{are} \xreplaced{conducted}{done} on the same public datasets to obtain the models’ performance. \xadded{These datasets were the classical datasets that are commonly exploited in KT research. }\xreplaced{Descriptions of the datasets are shown in}{Datasets are described as} Table~\ref{tab1}.

\begin{table}[H] 
\caption{Introduction of the datasets.\label{tab1}}
\newcolumntype{C}{>{\centering\arraybackslash}X}
\begin{tabularx}{\textwidth}{CCCC}
\toprule
\textbf{Dataset}	& \textbf{Students}	& \textbf{Skills} & \textbf{Records}\\
\midrule
ASSISTment2009 & 4151     & 110    & 325,637  \\
ASSISTment2015 & 19,840    & 100    & 683,801  \\
Synthetic-5    & 4000     & 50     & 200,000  \\
ASSISTchall    & 1709     & 102    & 942,816  \\
Statics2011    & 333      & 1223   & 189,297  \\
\bottomrule
\end{tabularx}
\end{table}
%\unskip

The ASSISTment2009 dataset comes from the log data of students performing math exercises in 2009, which \xreplaced{were}{is} collected by the ASSISTments platform~\cite{feng_addressing_2009}. The original version of the dataset \xreplaced{contained}{contains} 123 skills.
\xadded{Skills corresponded to exercise tags, such as prime numbers or linear equations in math problems. For the convenience of representation, the exercise tags were mapped to some numbers in the dataset. In other words, a skill ID represented an exercise tag in these KT datasets.}
Later, it \xreplaced{was}{is} found that there \xreplaced{were}{are} duplicate records of the data, which \xreplaced{affected}{affect} the reliability of the prediction results of the DKT model~\cite{xiong_going_2016}. Therefore, experiments in this paper \xreplaced{adopted}{adopt} the version after removing duplicate records. The total number of skills involved in this version of the dataset has been reduced to 110.

The ASSISTment2015 dataset comes from the data collected by the ASSISTments platform in 2015.

The Synthetic-5 dataset is from the paper which  \xreplaced{proposed}{proposes} DKT. The dataset is a set of unreal datasets constructed for experiments and does not correspond to the actual skills~\cite{piech_deep_2015}. However, the dataset is of good quality and is still suitable for many experimental studies in KT founded on DL.

The ASSISTchall dataset \xreplaced{was}{is} gathered when learners \xreplaced{utilized}{utilize} the ASSISTments platform and \xreplaced{was}{is} used for an educational data mining competition in 2017.

The Statics2011 dataset originates from {online college-level statistics lessons~\cite{koedinger_data_2010}.} %Please check that the intended meaning is retained.
% We confirm.

\xadded{When using the datasets mentioned above, the student \xreplaced{ whose records are less than or equal to}{ records with practice times less than or equal to} two were regarded as invalid records and were removed from the sample sets, as the student records with too few practice records could not reflect the real learning situation of the student and were not applicable to the time-series model~\cite{zhang_incorporating_2017}. }
\xadded{There were various situations leading to a student only practicing twice. The student may have already mastered a particular skill, and thus did not need practice. The student may also have given up the exercise because it was too difficult. So, the student records with two practice records did not seem reliable enough to reflect students' real learning situation. The training of neural networks relies on reliable samples on a large scale to obtain accurate results. So, we removed the students with less than two practice records, which could affect the whole model. }

\begin{figure}[H]
\vspace{-4pt}
%\begin{adjustwidth}{-\extralength}{0cm}
%\centering
\includegraphics[width=13.5cm]{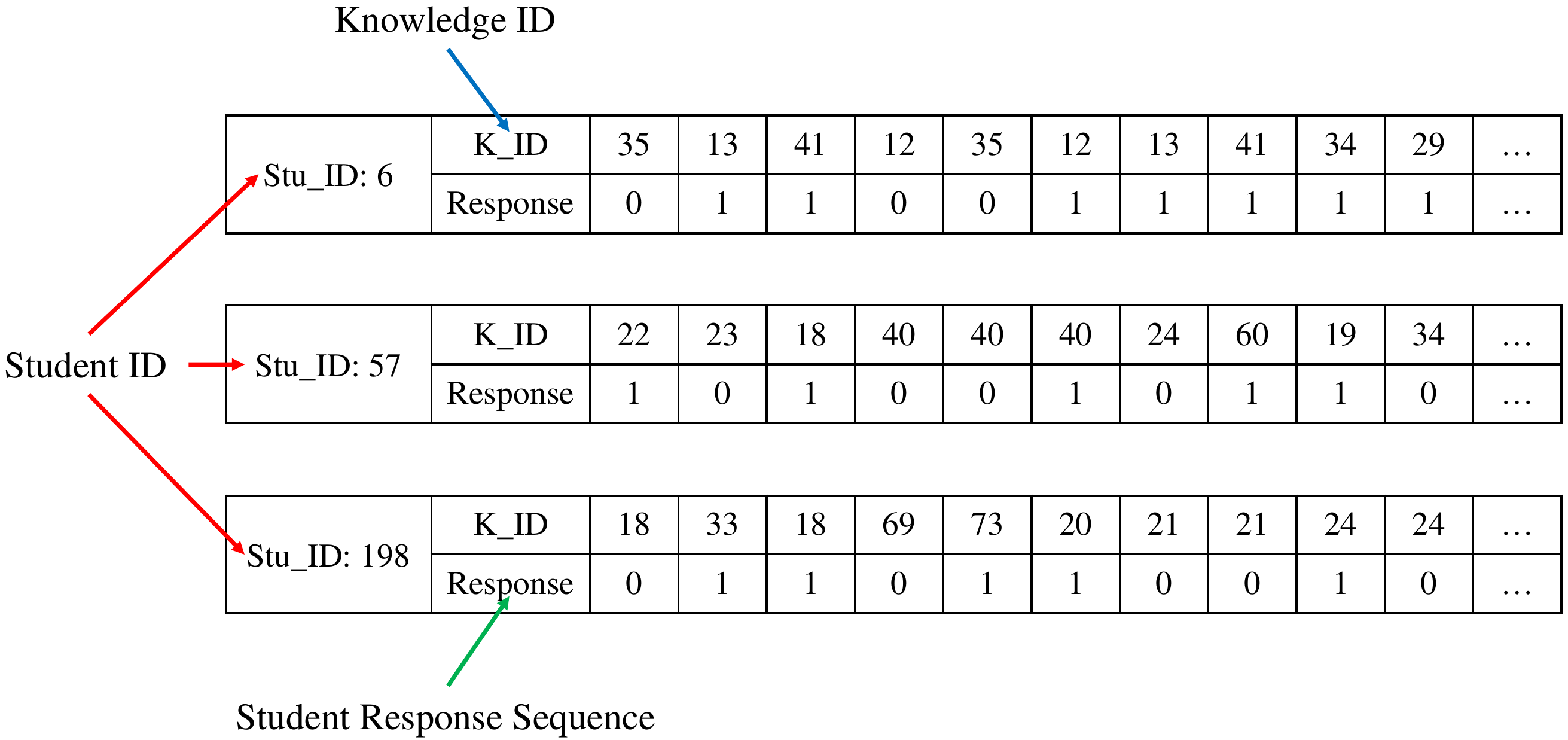}
%\end{adjustwidth}
\caption{\xadded{Description of a portion of the dataset. 
% Starting from the first line, each 3 lines of data are taken as a group data which represents the exercise records of a single student. The first row of each group represents the number of times the student has practiced, the second row represents the sequence of skill tags, and the third row represents the sequence of student answer correctness.  The number of times the student has exercised is equal to the length of the skill tag sequence, and also equal to the length of the correctness sequence.
} \label{fig7}}
\end{figure}

\xadded{A portion of the dataset is shown in Figure~\ref{fig7}.} 

% %\unskip

\subsection{{Experimental Environment}}
\xadded{
The experiments for \xreplaced{analyzing}{analysing} the models cannot be \xreplaced{completed}{done} without the support of an effective software and hardware environment. \xreplaced{A description of the main hardware and software environment configurations is shown in}{The main hardware and software environment configurations used in our experiments are described as} Table~\ref{tab5}. 
}

\begin{table}[H] 
\caption{{Description of the experimental environment.\label{tab5}}}
\newcolumntype{C}{>{\centering\arraybackslash}X}
{
\begin{tabularx}{\textwidth}{CC}
\toprule
\textbf{Configuration Environment}	& \textbf{Configuration Parameters} \\
\midrule
Operating System                   & Windows 10 64-bit                       \\
GPU                                & gtx 1080ti       \\
CPU                                & E5 Series (4 cores)                     \\
Memory                             & 16 GB                                   \\
Programming language               & Python 3.6                              \\
Deep learning framework            & Tensorflow 1.5                          \\
Python library                     & Numpy, Scikit-learn, Pandas, Matplotlib \\
\bottomrule
\end{tabularx}
}
\end{table}
% %\unskip

\subsection{Results and Discussion}
During our experiments, we \xreplaced{partitioned}{partition} each dataset into three parts: a training set, a validation set, and a test set. They, respectively, \xreplaced{accounted}{account} for about $55\%$, $15\%$, and $30\%$ of the total students. The training set \xreplaced{was}{is} used for training the model. The validation set \xreplaced{was}{is} used for adjusting hyperparameters. The test set \xreplaced{was}{is} used for evaluating the model. To obtain more reliable results, the experiments of each model on each dataset \xreplaced{were}{are} repeated three times, and the average values \xreplaced{were}{are} taken as the evaluation result in terms of RMSE, AUC, ACC, and $r^{2}$. Considering the balance of the calculation resource and the prediction accuracy, the hyperparameters \xreplaced{were}{are} set as in Table~\ref{tab2}. In fact, the DKT-STDRL model can obtain more accurate predictions by changing the hyperparameters, regardless of \mbox{calculation resources}.

\begin{table}[H] 
\caption{Hyperparameters of the DKT-STDRL.\label{tab2}}
\newcolumntype{C}{>{\centering\arraybackslash}X}
\begin{tabularx}{\textwidth}{CC}
\toprule
\textbf{Hyperparameters}	& \textbf{Value} \\
\midrule
Learning rate                                   & 0.001        \\
Rate of decay for the learning rate             & 0.3          \\
Steps before the decay                          & 8            \\
Batch size                                      & 50           \\
Epochs                                          & 10           \\
Shape of filters of the conv1d                  & (16, 50, 50) \\
Layers of the hierarchical convolution          & 3            \\
Keep probability for dropout of the convolution & 0.2          \\
Number of units of each LSTM cell               & 30           \\
Output keep probability of LSTM cell            & 0.3          \\
\bottomrule
\end{tabularx}
\end{table}
% %\unskip

\textbf{{Baselines comparison.} %MDPI: Please confirm if the bold should be retained. If not, please remove it.
%Authors: We confirm the bold should be retained.
} For the sake of better verifying the validity of the new model, DKT-STDRL \xreplaced{was}{is} compared with DKT based on LSTM and CKT on the same dataset. As the starting model of deep knowledge tracing, the DKT model based on LSTM had \xadded{an }important reference value. In addition, because DKT-STDRL is an improvement based on the CKT model, the comparison with the CKT model was essential. We refer to and reproduce the code of DKT and CKT as baselines, and obtain the evaluation results approach to the reported results.

Table~\ref{tab3} compares the prediction effects of DKT-STDRL with DKT and CKT. It was found in the experiments on the five datasets that our DKT-STDRL outperforms the DKT and the CKT in terms of RMSE, AUC, ACC, and $r^{2}$. Specifically, for ASSISTments2009, DKT-STDRL \xreplaced{achieved}{achieves} an RMSE of 0.2826. It \xreplaced{decreased}{decreases} by $11.8\%$ and $10.86\%$ compared, respectively, with DKT and CKT. DKT-STDRL \xreplaced{achieved}{achieves} an AUC of 0.9591. It \xreplaced{increased}{increases} by $15.42\%$ and $13.54\%$ compared, respectively, with DKT and CKT. DKT-STDRL \xreplaced{achieved}{achieves} an ACC of 0.8904. It \xreplaced{increased}{increases} by 12.41$\%$ and 11.56$\%$ compared, respectively, with DKT and CKT. DKT-STDRL \xreplaced{achieved}{achieves} $r^{2}$ of 0.644. It \xreplaced{increased}{increases} by 36.05$\%$ and 32.59$\%$ comparedm respectively, with DKT and CKT. For ASSISTments2015, Synthetic-5, ASSISTchal, and Statics2011, DKT-STDRL \xreplaced{had}{has} similar performance. Experiments verify the prediction effect of the model from two aspects: classification and regression. AUC and ACC of DKT-STDRL \xreplaced{were}{are} gained. Therefore, the DKT-STDRL model can better classify the prediction results of whether students can answer correctly or not. Moreover, from the perspective of regression, the reduction of RMSE with DKT-STDRL reflects that the new model is more accurate in predicting the probability of students answering correctly or incorrectly. In addition, the $r^{2}$ of DKT-STDRL has been significantly improved, which shows that the prediction results of the new model are highly correlated with the actual exercise performance of students. The new model learns the essential law of the change of students’ knowledge state from the sample data. Therefore, the DKT-STDRL model promotes present KT models predicting more accurately.

\begin{table}[H] 
\caption{Comparison of experimentation results among DKT-STDRL, DKT, and CKT models.\label{tab3}}
\newcolumntype{C}{>{\centering\arraybackslash}X}
\begin{tabularx}{\textwidth}{CCCCCC}
\toprule
\textbf{Datasets}	& \textbf{Models} & \textbf{RMSE} & \textbf{AUC} & \textbf{ACC} & \textbf{$r^{2}$}\\
\midrule
\multirow{3}{*}{ASSISTment2009} & \textbf{{Ours} %MDPI: Please confirm if the bold in this table should be retained. If so, please check whether need explanation for it. If not, please remove it. Same meaning as table 6.
%Authors: We confirm the bold in this table should be retained and it means the best performance method, and we already given the explanation in the manuscript.  
}  & \textbf{{0.2826}}  & \textbf{{0.9591}}  & \textbf{{0.8904}}  & \textbf{{0.6440}} \\
                                & DKT    & 0.4006 & 0.8049 & 0.7663 & 0.2835 \\
                                & CKT    & 0.3912 & 0.8237 & 0.7748 & 0.3181 \\ \midrule
\multirow{3}{*}{ASSISTment2015} & \textbf{Ours}  & \textbf{0.0766} & \textbf{0.9996}   & \textbf{0.9948}   & \textbf{0.9698} \\
                                & DKT    & 0.4131 & 0.7235 & 0.7504 & 0.1235 \\
                                & CKT    & 0.4107 & 0.7322 & 0.7542 & 0.1338 \\ \midrule
\multirow{3}{*}{Synthetic-5}    & \textbf{Ours}  & \textbf{0.3167} & \textbf{0.9482}   & \textbf{0.8790}   & \textbf{0.5766} \\
                                & DKT    & 0.4109 & 0.8167 & 0.7475 & 0.2868 \\
                                & CKT    & 0.4051 & 0.8279 & 0.7553 & 0.3075 \\ \midrule
\multirow{3}{*}{ASSISTchall}    & \textbf{Ours}  & \textbf{0.2716} & \textbf{0.9710}   & \textbf{0.9078}   & \textbf{0.6847}  \\
                                & DKT    & 0.4538 & 0.7022 & 0.6791 & 0.1198 \\
                                & CKT    & 0.4500 & 0.7127 & 0.6857 & 0.1342 \\ \midrule
\multirow{3}{*}{Statics2011}    & \textbf{Ours}  & \textbf{0.2772} & \textbf{0.9499}   & \textbf{0.9010}   & \textbf{0.5621}  \\
                                & DKT    & 0.3697 & 0.8012 & 0.8054 & 0.2216 \\
                                & CKT    & 0.3630 & 0.8232 & 0.8101 & 0.2496          \\
\bottomrule
\end{tabularx}
\end{table}

\textbf{{Comparison of the variants of the DKT-STDRL.}} The experiments for comparing DKT-STDRL with its variants, by removing different parts, are helpful to understand the importance of various components in the DKT-STDRL model for advancing the prediction effect of the whole model. Therefore, in addition to the CKT and DKT, four variants \xreplaced{were}{are} designed by us to observe the different impacts on the prediction effect from the aspects of spatial features, temporal features, prior features, and joint features. The four variants are described as follows:
\begin{itemize}
\item    \emph{{DKT-TDRL.} %MDPI: Please confirm if the italics should be retained.
%Authors: We confirm the italics should be retained.
} To study the influence of spatial features on the prediction effect of the DKT-STDRL model, we removed the part of extracting spatial features with CNN from the DKT-STDRL model and obtaine dthe variant model DKT-TDRL. Specifically, the DKT-TDRL model first takes the input data represented by the embedding matrix and the students’ personalized prior knowledge state as the prior feature. Then, through the intermediate data processing process, it is input into BiLSTM for bidirectional time feature extraction. Finally, the prediction result is output;
\item	\emph{{DKT-SDRL1.}} For the purpose of studying the impact of time features of DKT-STDRL on the prediction effect, the part of extracting time features was removed from the DKT-STDRL model. Since DKT-STDRL adopts BiLSTM to the express bidirectional feature of the sequence, two schemes can be obtained to study the influence of unidirectional and bidirectional temporal feature extraction, respectively. The first scheme is to remove the bidirectional temporal feature extraction structure of the DKT-STDRL model. Following this idea, to obtain the prediction results of the next time step, the model is transformed into the CKT model. The second scheme is to change the BiLSTM structure of the DKT-STDRL model into \xadded{the} LSTM structure, which can show the influence of one-way temporal output on the prediction effect of the model. Then, we can compare the difference between two-way time characteristics and one-way time characteristics in solving the prediction problem of students’ learning performance. We abbreviate the second scheme as DKT-SDRL1. Specifically, DKT-SDRL1 first extracts and uses the prior learning features, and then a multi-layer convolution structure is used to extract students’ spatial learning features. Then, after a simple intermediate process, it is input to the one-way temporal feature extraction layer based on LSTM. Finally, the prediction results are output;
\item	\emph{{DKT-STDRRP.}} In order to study the influence of the prior learning features of the DKT-STDRL model on the prediction effect, we removed the part of extracting the prior learning features from the DKT-STDRL model. This variant model is called DKT-STDRRP. DKT-STDRRP transforms the input students’ learning history data into an embedded matrix, and then extracts spatial features through CNN layers and extracts bidirectional temporal features through the BiLSTM layer, so as to predict students’ learning performance. By comparing the prediction results before and after removing the prior learning features, it is helpful to intuitively understand the role of the prior features in the task of predicting learning performance;
\item	\emph{{DKT-STDRRJ.}} For studying the impact of the joint features in the DKT-STDRL model on the prediction, the process of merging the one-hot coded exercise history data is removed from the DKT-STDRL model in the intermediate data processing. In other words, the variant scheme keeps the structure of the first half of the DKT-STDRL model unchanged, inputs the one-hot coded spatial features directly into the temporal feature extraction part, and then obtains the final prediction result. This scheme is called DKT-STDRRJ.
\end{itemize}

Table~\ref{tab4} shows the prediction performance of DKT-STDRL, CKT, DKT, and the variants of DKT-STDRL (DKT-TDRL, DKT-SDRL1, DKT-STDRRP, and DKT-STDRRJ) on the ASSISTment2009, ASSISTment2015, Synthetic-5, ASSISTchall, and Statics2011 datasets. It should be noted that different models are based on the same hyperparameters in the experiments on the same datasets. Because as parts of the DKT-STDRL, other models should keep the same settings {to make only the model structure different in each experiment.} %Please check that the intended meaning is retained. 
% We confirm.

\begin{figure}[H]
\includegraphics[width=0.8\textwidth]{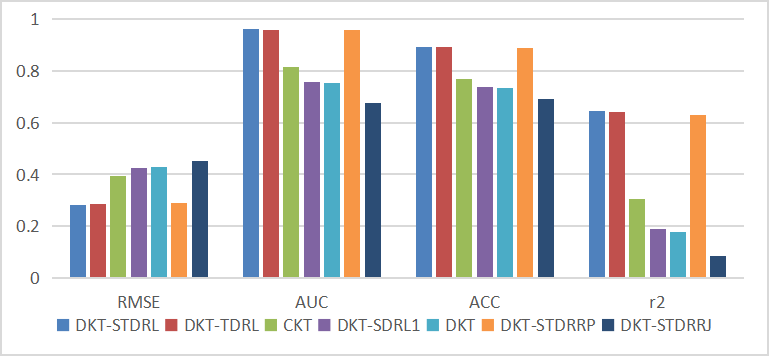}
\caption{Prediction results of DKT-STDRL, CKT, DKT, and the variants of DKT-STDRL on the ASSISTments2009 dataset.\label{fig2}}
\end{figure}

\begin{figure}[H]
\includegraphics[width=0.8\textwidth]{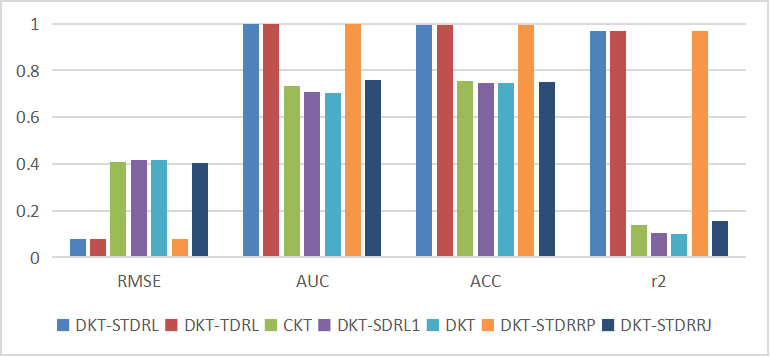}
\caption{Prediction results of DKT-STDRL, CKT, DKT, and the variants of DKT-STDRL on the ASSISTments2015 dataset.\label{fig3}}
\end{figure}

\begin{figure}[H]
\includegraphics[width=0.8\textwidth]{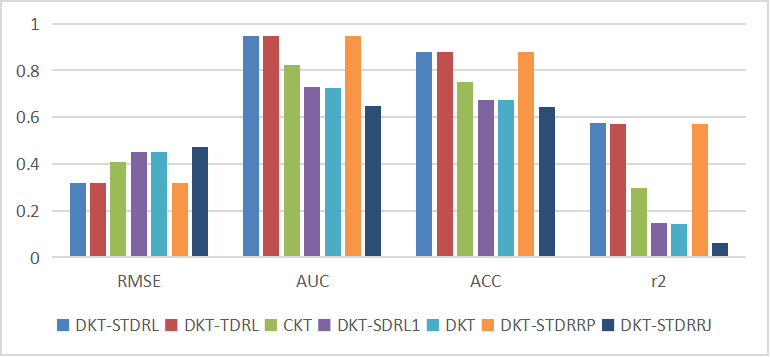}
\caption{Prediction results of DKT-STDRL, CKT, DKT, and the variants of DKT-STDRL on the Synthetic-5~dataset.\label{fig4}}
\end{figure}

\begin{figure}[H]
\includegraphics[width=0.8\textwidth]{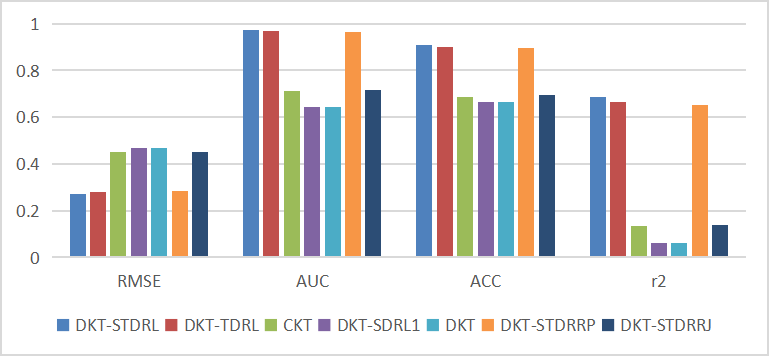}
\caption{Prediction results of DKT-STDRL, CKT, DKT, and the variants of DKT-STDRL on the ASSISTchall~dataset.\label{fig5}}
\end{figure}

\begin{figure}[H]
\includegraphics[width=0.8\textwidth]{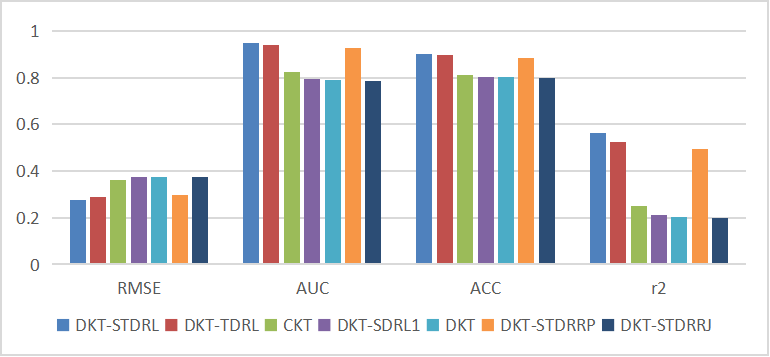}
\caption{Prediction results of DKT-STDRL, CKT, DKT, and the variants of DKT-STDRL on the Statics2011~dataset.\label{fig6}}
\end{figure}

In order to more intuitively compare the prediction effect of these models, \xreplaced{Figure~\ref{fig2} displays the prediction results of all models on the ASSISTment2009 dataset in the bar charts.
Figure~\ref{fig3} displays the prediction results of all models on the ASSISTment2015 dataset in the bar charts.
Figure~\ref{fig4} displays the prediction results of all models on the Synthetic-5 dataset in the bar charts.
Figure~\ref{fig5} displays the prediction results of all models on the ASSISTchall dataset in the bar charts.
Figure~\ref{fig6} displays the prediction results of all models on the Statics2011 dataset in the bar charts.}
{Figure~\ref{fig2}, Figure~\ref{fig3}, Figure~\ref{fig4}, Figure~\ref{fig5}, and Figure~\ref{fig6} displays the prediction effects of all models in the bar charts.} 
\xadded{From these figures, it can be seen, obviously, that the DKT-STDRL had a higher AUC, ACC\xadded{,} and $r^{2}$, and a lower RMSE, than CKT, DKT, DKT-SDRL1\xadded{,} and DKT-STDRRJ. Moreover, the AUC, ACC\xadded{,} and $r^{2}$ of the DKT-STDRL were a little higher than DKT-TDRL and DKT-STDRRP, and the RMSE of DKT-STDRL was a little lower than DKT-TDRL and DKT-STDRRP. The model with higher AUC, ACC\xadded{,} and $r^{2}$, and lower RMSE, is better. }
\xreplaced{So, }{It can be seen that }DKT-STDRL is significantly better than CKT, DKT, DKT-SDRL1\xadded{,} and DKT-STDRRJ. Moreover, DKT-STDRL is only a little better than DKT-TDRL and DKT-STDRRP.  Firstly, DKT-TDRL, regardless of extracting spatial features, \xreplaced{raised}{raises} the RMSE no more than 2$\%$ and \xreplaced{dropped}{drops} the AUC, ACC, $r^{2}$ no more than 2$\%$, 0.7$\%$, and 5$\%$, respectively. So, the spatial features have \xdeleted{a }little benefit in improving prediction results. Secondly, different from DKT-STDRL, DKT-SDRL1 learns unidirectional temporal features instead of bidirectional features. As the results show, DKT-SDRL1 is significantly worse than DKT-STDRL. Thus, the extracted bidirectional temporal features are very important for DKT-STDRL. Thirdly, DKT-STDRL has many advantages compared with DKT-STDRRP. Changing from DKT-STDRRP to DKT-STDRL can decrease the RMSE \xreplaced{by}{of} over 2$\%$ and increase AUC, ACC, and $r^{2}$ by over 2$\%$, 1.8$\%$, 6.8$\%${, repsectively}. %Please check that the intended meaning is retained. 
So, prior learning features are also meaningful for obtaining a better model. Fourthly, DKT-STDRRJ has obvious weaknesses in all the metrics. It can be inferred that merging exercise performance history information with \xdeleted{the }spatial features is necessary for the DKT-STDRL because the extracted spatial features have information errors compared with the original data and cannot be directly input into BiLSTM. So, the spatial features, temporal features, prior features, and joint features of DKT-STDRL play various roles in improving prediction.

\begin{table}[H]
\caption{Comparison of experimentation results among the DKT-STDRL, CKT, DKT, and variants \mbox{of DKT-STDRL}. \label{tab4}}
	\begin{adjustwidth}{-\extralength}{0cm}
		\newcolumntype{C}{>{\centering\arraybackslash}X}
		\begin{tabularx}{\fulllength}{CCCCCC}
			\toprule
\textbf{Datasets}	& \textbf{Models} & \textbf{RMSE} & \textbf{AUC} & \textbf{ACC} & \textbf{$r^{2}$}\\
\midrule
\multirow{7}{*}{ASSISTment2009} & \textbf{{DKT-STDRL} %MDPI: Please confirm if the bold in this table should be retained.
%Authors: We confirm that the bold in this table should be retained.
}  & \textbf{{0.2826}}  & \textbf{{0.9591}}  & \textbf{{0.8904}}  & \textbf{{0.6440}} \\
                                & DKT-TDRL   & 0.2845 & 0.9574 & 0.8896 & 0.6394 \\
                                & CKT        & 0.3953 & 0.8154 & 0.7687 & 0.3039 \\
                                & DKT-SDRL1  & 0.4265 & 0.7567 & 0.7370 & 0.1894 \\
                                & DKT        & 0.4293 & 0.7509 & 0.7338 & 0.1788 \\
                                & DKT-STDRRP & 0.2878 & 0.9557 & 0.8861 & 0.6310 \\
                                & DKT-STDRRJ & 0.4531 & 0.6736 & 0.6890 & 0.0853 \\ \midrule
\multirow{7}{*}{ASSISTment2015} & \textbf{DKT-STDRL}  & \textbf{0.0766} & \textbf{0.9996}   & \textbf{0.9948}   & \textbf{0.9698} \\
                                & DKT-TDRL   & 0.0799 & 0.9995 & 0.9942 & 0.9672 \\
                                & CKT        & 0.4099 & 0.7343 & 0.7546 & 0.1370 \\
                                & DKT-SDRL1  & 0.4176 & 0.7070 & 0.7475 & 0.1043 \\
                                & DKT        & 0.4183 & 0.7042 & 0.7468 & 0.1015 \\
                                & DKT-STDRRP & 0.0801 & 0.9995 & 0.9943 & 0.9671 \\
                                & DKT-STDRRJ & 0.4053 & 0.7581 & 0.7511 & 0.1563 \\ \midrule
\multirow{7}{*}{Synthetic-5}    & \textbf{DKT-STDRL}  & \textbf{0.3167} & \textbf{0.9482}   & \textbf{0.8790}   & \textbf{0.5766} \\
                                & DKT-TDRL   & 0.3186 & 0.9469 & 0.8776 & 0.5714 \\
                                & CKT        & 0.4081 & 0.8241 & 0.7513 & 0.2972 \\
                                & DKT-SDRL1  & 0.4496 & 0.7283 & 0.6754 & 0.1469 \\
                                & DKT        & 0.4505 & 0.7269 & 0.6752 & 0.1432 \\
                                & DKT-STDRRP & 0.3194 & 0.9461 & 0.8768 & 0.5693 \\
                                & DKT-STDRRJ & 0.4716 & 0.6485 & 0.6417 & 0.0612 \\ \midrule
\multirow{7}{*}{ASSISTchall}    & \textbf{DKT-STDRL}  & \textbf{0.2716} & \textbf{0.9710}   & \textbf{0.9078}   & \textbf{0.6847}  \\
                                & DKT-TDRL   & 0.2790 & 0.9675 & 0.9009 & 0.6671 \\
                                & CKT        & 0.4503 & 0.7119 & 0.6846 & 0.1330 \\
                                & DKT-SDRL1  & 0.4683 & 0.6431 & 0.6642 & 0.0627 \\
                                & DKT        & 0.4682 & 0.6430 & 0.6648 & 0.0628 \\
                                & DKT-STDRRP & 0.2854 & 0.9633 & 0.8942 & 0.6518 \\
                                & DKT-STDRRJ & 0.4493 & 0.7170 & 0.6935 & 0.1371 \\ \midrule
\multirow{7}{*}{Statics2011}    & \textbf{DKT-STDRL}  & \textbf{0.2772} & \textbf{0.9499}   & \textbf{0.9010}   & \textbf{0.5621}  \\
                                & DKT-TDRL   & 0.2891 & 0.9395 & 0.8956 & 0.5220 \\
                                & CKT        & 0.3630 & 0.8241 & 0.8099 & 0.2496 \\
                                & DKT-SDRL1  & 0.3724 & 0.7942 & 0.8042 & 0.2103 \\
                                & DKT        & 0.3739 & 0.7891 & 0.8023 & 0.2036 \\
                                & DKT-STDRRP & 0.2974 & 0.9275 & 0.8827 & 0.4940 \\
                                & DKT-STDRRJ & 0.3756 & 0.7873 & 0.7995 & 0.1967          \\
\bottomrule
		\end{tabularx}
	\end{adjustwidth}
\end{table}

% Bulleted lists look like this:
% \begin{itemize}
% \item	First bullet;
% \item	Second bullet;
% \item	Third bullet.
% \end{itemize}

% Numbered lists can be added as follows:
% \begin{enumerate}
% \item	First item; 
% \item	Second item;
% \item	Third item.
% \end{enumerate}

\vspace{-9pt}
% %\unskip
\vspace{-6pt}
% %\unskip
\vspace{-6pt}
% %\unskip
\vspace{-6pt}
% %\unskip
\xadded{
From the previous two types of comparison experiments, it can be found that our \mbox{ DKT-STDRL} has better prediction accuracy than CKT~\cite{shen_convolutional_2020} and DKT~\cite{piech_deep_2015}. 
Analyzing the functions of the various parts of DKT-STDRL, the spatial features, temporal features, prior features, and joint features of DKT-STDRL all contribute to improving the prediction accuracy to varying degrees.  In short, the success of DKT-STDRL in improving the prediction accuracy is inseparable from the comprehensive analysis of the spatial and temporal characteristics of students' practice sequences.  In other words, for KT, a comprehensive consideration of students' personalized learning efficiency and characteristics of students' knowledge state change process is helpful to obtain more accurate prediction results.  
}

% %\unskip

\xadded{ 
\xreplaced{Our model significantly improved the prediction accuracy of KT, }{The prediction accuracy of \xadded{the }DKT-STDRL model is significantly improved, }
and it had practical significance for \xreplaced{different education stakeholders, such as students, LMS administrators, teachers, and instructional designers~\cite{cavus_distance_2015}}{the development of personalized education based on LMS}.  First, \xreplaced{LMSs which use our method}{LMS} can analyze students' potential knowledge state based on the students' practice logs and make \xadded{an }accurate prediction of students' performance in the next stage\xdeleted{by using our DKT-STDRL}\xreplaced{. Therefore, our method can help the LMSs to obtain information about students' future learning performance.  LMSs can generate reports for students or teachers to help system users better find the problems in learning. LMSs can also learn the knowledge points that students have not mastered from the prediction information of students' learning performance, so as to recommend content suitable for students' further learning. }
{, so that learning resources suitable for students can be intelligently recommended.} 
\xreplaced{In}{By} this way, \xreplaced{LMSs}{the LMS} can send exercises that students really need without wasting too much time \xdeleted{in} doing exercises they have grasped. Secondly, \xreplaced{LMSs}{the LMS system} using our KT model can provide teachers and instructional designers with more accurate analysis reports on students' learning conditions based on the prediction of students' answer performance, so as to help teachers and instructional designers flexibly adjust teaching plans. \xadded{Through the learning reports provided by the systems, teachers and instructional designers can prepare lessons efficiently around students' knowledge defects. } \xreplaced{With}{By} the continuous improvement of teaching methods, \xadded{the }student-centered outcome-based education (OBE) concept can be realized. Thirdly, the administrators of \xreplaced{LMSs}{LMS} adopting our KT model can provide better consulting services for customers. \xadded{Administrators can also divide classes according to the degree of knowledge mastery of students and assign special teachers to the classes for guidance, which is convenient for class management and improves the overall teaching effect for students. } So, our KT model can promote the development of intelligent education.
}

\section{Conclusions}\label{sec5}
For filling the gap of current KT models that express students’ learning features insufficiently, Deep Knowledge Tracing Based on Spatial and Temporal Deep Representation Learning for Learning Performance Prediction (DKT-STDRL) \xreplaced{was}{is} put forward. \mbox{DKT-STDRL } \xreplaced{extracted}{extracts} the spatial features on the basis of students’ exercise history, and then further \xreplaced{extracted}{extracts} the temporal features of students’ exercise sequence. Firstly,\linebreak \mbox{DKT-STDRL} \xreplaced{used}{uses} CNN to extract the spatial feature information of students’ exercise history. Then, the spatial features \xreplaced{were}{are} connected with the exercise history features and input into the BiLSTM part as joint learning features. Finally, BiLSTM \xreplaced{extracted}{extracts} the temporal feature from the joint learning features to obtain the prediction information of whether the students \xreplaced{could}{can} answer correctly or not at the next time step. The prediction effect of the DKT-STDRL model \xreplaced{was}{is} verified on five common public datasets. The experimentations \xreplaced{demonstrated}{demonstrate} that the prediction of the DKT-STDRL \xreplaced{outperformed}{outperforms} the DKT and CKT. Therefore, DKT-STDRL is effective for promoting the prediction accuracy of the KT model on the basis of DL. Moreover, many experiments  \xreplaced{were conducted}{are done} to compare the prediction performance of DKT-STDRL with CKT, DKT, and four variants of DKT-STDRL, which \xreplaced{showed}{show} the different impacts on the prediction effect from the aspects of spatial features, temporal features, prior features, and joint features.

\xreplaced{Though we have \xreplaced{succeeded}{succeed} in advancing the prediction accuracy, there are still some limitations of our work on the complexity and interpretability of the model. First, the structure of the model is complex\xreplaced{ and}{, which} has too many parameters. Secondly, the parameters in the DL networks lack  \xdeleted{of the }interpretability, which limits the significance of the model in practical applications. So, in prospective work, we strive to reduce the complexity of the model to save the computing resources and to improve the interpretability to provide more application value.}{In prospective work, reducing the complexity of the model can be tried to get better performance and interpretability.} In addition, \xadded{in the future,} because the model has good prediction accuracy, we can try to integrate the model into \xreplaced{LMSs}{online education platforms} for \xadded{a }better recommendation, so that we obtain more intelligent and caring online learning systems. \xadded{By using our improved KT algorithm, LMSs can obtain more accurate information about students' knowledge mastery or prediction results of answer performance. Based on this information, LMSs can intelligently recommend learning resources (such as lectures, documents, exercises, and quizzes) that better meet students' needs, thus helping students reduce their learning burden and improve their learning efficiency.}

%%%%%%%%%%%%%%%%%%%%%%%%%%%%%%%%%%%%%%%%%%
% \section{Discussion}

% Authors should discuss the results and how they can be interpreted from the perspective of previous studies and of the working hypotheses. The findings and their implications should be discussed in the broadest context possible. Future research directions may also be highlighted.

%%%%%%%%%%%%%%%%%%%%%%%%%%%%%%%%%%%%%%%%%%
% \section{Conclusions}

% This section is not mandatory, but can be added to the manuscript if the discussion is unusually long or complex.

%%%%%%%%%%%%%%%%%%%%%%%%%%%%%%%%%%%%%%%%%%

\vspace{6pt} 

%%%%%%%%%%%%%%%%%%%%%%%%%%%%%%%%%%%%%%%%%%
%% optional
%\supplementary{The following supporting information can be downloaded at:  \linksupplementary{s1}, Figure S1: title; Table S1: title; Video S1: title.}

% Only for the journal Methods and Protocols:
% If you wish to submit a video article, please do so with any other supplementary material.
% \supplementary{The following supporting information can be downloaded at: \linksupplementary{s1}, Figure S1: title; Table S1: title; Video S1: title. A supporting video article is available at doi: link.}

%%%%%%%%%%%%%%%%%%%%%%%%%%%%%%%%%%%%%%%%%%
\authorcontributions{Methodology research, L.L. and Z.W.; model realization, L.L.; supervision Z.W.; writing and editing, L.L. and Z.W.; data collection, L.L., Z.W., H.Y., Z.Y. and Y.L.; model evaluation, L.L., Z.W., H.Y., Z.Y. and Y.L.; funding acquisition, L.L., Z.W., H.Y., Z.Y. and Y.L. All authors have read and agreed to the published version of the manuscript.}

\funding{ This research was funded in part by the Youth Science Foundation of Heilongjiang Institute of Technology (2021QJ07), the National Natural Science Foundation of China (Nos. 62177022, 61901165, 61501199), the Collaborative Innovation Center for Informatization and Balanced Development of K-12 Education by MOE and Hubei Province (No. xtzd2021-005), and Self-determined Research Funds of CCNU from the Colleges’ Basic Research and Operation of MOE (No. CCNU20ZT010), the Natural Science Foundation of Heilongjiang Province (LH2020F047), the Innovation Team Project of Heilongjiang Institute of Technology (2020CX07), the University Nursing Program for Young Scholars with Creative Talents in Heilongjiang Province (UNPYSCT-2020052)\xadded{, and the Education and Teaching Reform Research Project of Heilongjiang Institute of Technology (JG202109)}. }

 \institutionalreview{{Not applicable}.%In this section, you should add the Institutional Review Board Statement and approval number, if relevant to your study. You might choose to exclude this statement if the study did not require ethical approval. Please note that the Editorial Office might ask you for further information. Please add “The study was conducted in accordance with the Declaration of Helsinki, and approved by the Institutional Review Board (or Ethics Committee) of NAME OF INSTITUTE (protocol code XXX and date of approval).” for studies involving humans. OR “The animal study protocol was approved by the Institutional Review Board (or Ethics Committee) of NAME OF INSTITUTE (protocol code XXX and date of approval).” for studies involving animals. OR “Ethical review and approval were waived for this study due to REASON (please provide a detailed justification).” OR “Not applicable” for studies not involving humans or animals.
}

 \informedconsent{{Not applicable}.%Any research article describing a study involving humans should contain this statement. Please add ``Informed consent was obtained from all subjects involved in the study.'' OR ``Patient consent was waived due to REASON (please provide a detailed justification).'' OR ``Not applicable'' for studies not involving humans. You might also choose to exclude this statement if the study did not involve humans.

% Written informed consent for publication must be obtained from participating patients who can be identified (including by the patients themselves). Please state ``Written informed consent has been obtained from the patient(s) to publish this paper'' if applicable.
}

\dataavailability{The ASSISTment2009, ASSISTment2015, and ASSISTchall datasets are from the ASSISTments platform~\cite{feng_addressing_2009}. Synthetic-5 is a simulated dataset used by the DKT model, which can be obtained through the deep knowledge tracing paper~\cite{piech_deep_2015}. The Statics2011 dataset originates from {online college-level statistics lessons~\cite{koedinger_data_2010}.} %Please check that the intended meaning is retained. 
\xadded{In the paper of CKT~\cite{shen_convolutional_2020}, these datasets have been organized and can be found by following the following {link}: %MDPI: Please add the access date (can NOT be later than the received date 1 June 2022).
%Authors: The access date is 30 May 2022.
 \url{https://github.com/bigdata-ustc/Convolutional-Knowledge-Tracing}; accessed on 30 May 2022. }
} 

% \acknowledgments{{XXX}.%In this section you can acknowledge any support given which is not covered by the author contribution or funding sections. This may include administrative and technical support, or donations in kind (e.g., materials used for experiments).
% }

\conflictsofinterest{The authors declare no conflict of interest.} 

%%%%%%%%%%%%%%%%%%%%%%%%%%%%%%%%%%%%%%%%%%
%% Optional
% \sampleavailability{Samples of the compounds ... are available from the authors.}

%% Only for journal Encyclopedia
%\entrylink{The Link to this entry published on the encyclopedia platform.}

\abbreviations{Abbreviations}{
The following abbreviations are used in this manuscript:\\

\noindent 
\begin{tabular}{@{}ll}
\xadded{LMS} &
\xadded{Learning Management System}\\
KT & Knowledge Tracing\\
DKT-STDRL & Deep Knowledge Tracing Based on Spatial and Temporal Deep Representation \\ 
    & Learning for Learning Performance Prediction\\
BKT & Bayesian Knowledge Tracing\\
HMM & Hidden Markov Model\\
\xadded{KT-IDEM} &
\xadded{Knowledge Tracing: Item Difficulty Effect Model}\\
\xadded{PC-BKT} &
\xadded{Personalized Clustered BKT}\\
RNN & Recurrent Neural Network\\
DL & Deep Learning\\
DKVMN & Dynamic Key-Value Memory Network\\
\end{tabular}

%\abbreviations{}{%
\noindent
\begin{tabular}{@{}ll}
MANN & Memory-Augmented Neural Network\\
CNN & Convolutional Neural Network\\
CKT & Convolutional Knowledge Tracing\\

GKT & Graph-based Knowledge Tracing\\
GNN & Graph Neural Network\\
BiLSTM~~~~~~~~~~~& Bidirectional Long Short-Term Memory\\
IRT & Item Response Theory\\
LSTM & Long Short-Term Memory\\
GLU & Gate Linear Unit\\
SKVMN & Sequential Key-Value Memory Networks \\
EERNN & Exercise-Enhanced Recurrent Neural Network\\
SAKT & Self-Attentive Knowledge Tracing\\
HRP & Historical Relevant Performance \\
CPC & Concept-wise Percent Correct \\
RMSE & Root Mean Squared Error\\
AUC & Area Under Curve\\
ACC & Accuracy \\
\xadded{OBE} &
\xadded{Outcome-Based Education}
\end{tabular}
}

%%%%%%%%%%%%%%%%%%%%%%%%%%%%%%%%%%%%%%%%%%
%% Optional
% \appendixtitles{no} % Leave argument "no" if all appendix headings stay EMPTY (then no dot is printed after "Appendix A"). If the appendix sections contain a heading then change the argument to "yes".
% \appendixstart
% \appendix
% \section[\appendixname~\thesection]{}
% \subsection[\appendixname~\thesubsection]{}
% The appendix is an optional section that can contain details and data supplemental to the main text---for example, explanations of experimental details that would disrupt the flow of the main text but nonetheless remain crucial to understanding and reproducing the research shown; figures of replicates for experiments of which representative data are shown in the main text can be added here if brief, or as Supplementary Data. Mathematical proofs of results not central to the paper can be added as an appendix.

% \begin{table}[H] 
% \caption{This is a table caption.\label{tab5}}
% \newcolumntype{C}{>{\centering\arraybackslash}X}
% \begin{tabularx}{\textwidth}{CCC}
% \toprule
% \textbf{Title 1}	& \textbf{Title 2}	& \textbf{Title 3}\\
% \midrule
% Entry 1		& Data			& Data\\
% Entry 2		& Data			& Data\\
% \bottomrule
% \end{tabularx}
% \end{table}

% \section[\appendixname~\thesection]{}
% All appendix sections must be cited in the main text. In the appendices, Figures, Tables, etc. should be labeled, starting with ``A''---e.g., Figure A1, Figure A2, etc.

%%%%%%%%%%%%%%%%%%%%%%%%%%%%%%%%%%%%%%%%%%
\begin{adjustwidth}{-\extralength}{0cm}
%\printendnotes[custom] % Un-comment to print a list of endnotes

\reftitle{References}

\end{adjustwidth}
\end{document}